\documentclass[5p]{elsarticle}

\usepackage{parskip}
\usepackage{amsmath,amsfonts,amssymb}
\usepackage{graphicx} 
\usepackage{algorithm}
\usepackage{algpseudocode}
\usepackage{booktabs}
\usepackage{svg}
\usepackage{caption}
\usepackage{tabularx}
\usepackage{booktabs} 
\usepackage{multirow} 
\usepackage{xcolor}
\usepackage{hyperref}
\usepackage{tikz}
\usetikzlibrary{patterns}
\usetikzlibrary{shapes.geometric}
\newcommand{\SG}[1]{{\color{red}#1}}

\begin{document}

\begin{frontmatter}

\title{Fast and explainable clustering in the~Manhattan~and Tanimoto distance}

\author[inst1]{Stefan G\"{u}ttel}
\author[inst1]{Kaustubh Roy}
\cortext[cor1]{Corresponding author: kaustubh.roy@manchester.ac.uk}
\address[inst1]{Department of Mathematics, University of Manchester, United Kingdom}

\begin{abstract}
The CLASSIX algorithm [X.~Chen \& S.~G\"{u}ttel, \emph{Pattern Recognition,} 150:110298, 2024] is a fast and explainable approach to data clustering. In its original form, this algorithm exploits the sorting of the data points by their first principal component to truncate the search for nearby data points, with nearness being defined in terms of the Euclidean distance. Here we extend CLASSIX to other distance metrics, including the Manhattan distance and the Tanimoto distance. Instead of principal components, we use an appropriate norm of the data vectors as the sorting criterion, combined with the triangle inequality for search termination. In the case of Tanimoto distance, a provably sharper intersection inequality is used to further boost the performance of the new algorithm. 
On a real-world chemical fingerprint benchmark, CLASSIX is about 30 times faster than the Taylor--Butina algorithm, and about 80 times faster than DBSCAN, while computing higher-quality clusters in both cases.
\end{abstract}


\end{frontmatter}

\date{\today}

\let\oldmarginpar\marginpar
\renewcommand\marginpar[1]{\-\oldmarginpar[\raggedleft\footnotesize #1]%
{\raggedright\footnotesize #1}}

\begin{section}{Introduction}\label{introduction}
Unsupervised clustering is a widely used machine learning technique employed to find patterns and similarity between elements in large datasets. Clustering involves grouping data points in a way such that points in the same cluster have similar properties and points belonging to separate clusters are dissimilar. Broadly, clustering is done using either distance or density-based approaches. The two most commonly used distance-based methods are $k$-means~\cite{macqueen1967multivariate} and $k$-medoids \cite{PARK20093336}, and popular density-based methods include mean shift \cite{1000236}, Quickshift \cite{10.1007/978-3-540-88693-8_52}, DBSCAN~\cite{10.5555/3001460.3001507}, OPTICS \cite{10.1145/304182.304187}, among others~\cite{https://doi.org/10.13140/2.1.4554.6887}.

Distance-based methods such as $k$-means  partition the data into
$k$~clusters by iteratively relocating the cluster centers and reassigning data points to their nearest center. These methods perform well for spherical and well-separated clusters. Density-based methods like mean shift and DBSCAN assume that the data appear in connected regions of high density. Mean shift performs clustering by iteratively shifting data points to nearby modes of the empirical density function computed by kernel density estimation. Quickshift uses Euclidean medoids as shifts. DBSCAN  progressively scans through the data  using a radius-based neighborhood search. The original implementation uses an R* tree data structure to perform range queries. These tree structures are not trivial to implement efficiently and can make the algorithm have a worst-case complexity of $O(n^2)$ in practice.

CLASSIX integrates key elements of both distance- and density-based clustering paradigms, with its original formulation~\cite{CG24a} limited to the Euclidean norm. As a first step, CLASSIX performs a greedy aggregation phase, assigning data points into groups centered around so-called starting points, which are points of the data that serve as approximate density estimators akin to the neighborhood centers in DBSCAN. However, in CLASSIX the groups are also reused to cluster the data, not just as estimators. CLASSIX’s two-stage approach also shares features with hierarchical clustering methods like HDBSCAN~\cite{10.1007/978-3-642-37456-2_14}. A key ingredient of the original CLASSIX method is an initial sorting of the data points along the first principal component of the data, combined with an early termination criterion for the neighborhood search.

In this work, we show that CLASSIX can be readily adapted to more general distance metrics, demonstrated by implementing the algorithm in the Manhattan (or $p=1$) norm~$\|\cdot\|_1$ and the Tanimoto distance. The Tanimoto distance (also known as Jaccard distance) is widely used for comparing binary data, making it particularly suitable for applications such as text mining, chemistry, and bioinformatics. It measures the similarity between two sets by dividing the cardinalities of the intersection of two sets by their union. For binary vectors, the Tanimoto distance is defined as
\begin{equation}
D_{\text{Tanimoto}}(u, v) := 1 - \frac{\|u \land v\|_1}{\|u \lor v\|_1} \in [0,1], 
\label{eq:tanimoto}
\end{equation}
where $u$ and $v$ are binary vectors representing the sets.
A popular way of representing structural features of chemical compounds is by first encoding them as molecular ﬁngerprints, such as MACCS structural key3 and extended-connectivity ﬁngerprint (ECFP), commonly known as the Morgan fingerprint \cite{doi:10.1021/c160017a018}. Searching for molecules with similar chemical and physical properties or comprising of similar reaction groups then becomes a task of finding compounds with similar Morgan fingerprints. The Tanimoto distance is the most widely-used metric for this task \cite{taylor-butina-clustering-1999,tanimoto_index-BRH}.

Several established clustering algorithms, including DBSCAN, HDBSCAN, and OPTICS, have been extended to support the Manhattan and Tanimoto distances and are available through highly optimized implementations in the \texttt{scikit-learn} library. In this work, we compare the computational performance and clustering quality of CLASSIX with these widely-used methods.

\paragraph{Contributions and paper overview}
This paper extends the CLASSIX clustering framework beyond the Euclidean norm and makes several main contributions:
(i) we formulate a norm-agnostic aggregation strategy based on sorting by a scalar score derived from $\|x_i\|$ and using reverse-triangle-inequality bounds for early search termination;
(ii) for the Manhattan norm we study the effect of data shifting (``orthant shifting'') and provide analysis that explains when score-based pruning is effective;
(iii) for binary fingerprints clustered in the Tanimoto distance we derive and exploit a sharper pruning criterion based on the Baldi intersection inequality, enabling fast exact neighborhood filtering in high dimensions.

The structure of the paper is as follows. 
Section~2 briefly reviews the existing CLASSIX method for the Euclidean norm. Section~3 introduces the new general norm-based extension and provides an efficiency analysis for the Manhattan norm. Section~4 develops the Tanimoto distance variant, including deterministic bounds and an average-case probabilistic efficiency analysis. Section~5 reports extensive experiments on synthetic and real datasets (including chemical fingerprints) comparing runtime and clustering quality against established baselines. Section~6 briefly discusses explainability. We conclude in Section~7 and provide an outlook on future work. Our code can be found on Github \footnote{\url{https://github.com/kaustubhroy1995/classix_tanimoto_manhattan}}.

\section{CLASSIX in the Euclidean distance}\label{classix:euclidean}

Let us briefly review the original CLASSIX method~\cite{CG24a} in order to put our contribution in context and introduce notation.

Assume we are given data points $x_1, x_2, \dots, x_n\in \mathbb{R}^d$ which are already mean-centered by subtracting the empirical mean of each feature from the vectors. An SVD is performed on the $n \times d$ data matrix $X := [x_1, x_2 , \dots, x_n]^T$, 
$$
    X = U \Sigma V^T
$$
where $U \in \mathbb R^{n\times d},   V \in \mathbb R^{d\times d}$  have orthonormal columns and $\Sigma \in \mathbb R^{d\times d}$ is a diagonal matrix with elements $\sigma_1 \geq \sigma_2 \geq \dots \geq \sigma_d\geq 0$. The data points are then scored according to  their first principal component~$\alpha_i$ given by
\begin{equation*}
    \alpha_i = {x_i}^T v_1 
\end{equation*}
where $v_1\in \mathbb{R}^d$ is the first column of $V$. 
The data is then sorted so that in the new ordering, $\alpha_1 \leq \alpha_2 \leq \dots \leq \alpha_n$.

In CLASSIX's aggregation phase, data points are grouped together based on the Euclidean distance criterion
\begin{equation*}
    D(u , v) := \| u - v \|_2 \leq \texttt{radius}. 
\end{equation*}
Here, \texttt{radius} is a hyperparameter chosen by the user. In order to reduce pairwise distance computations, CLASSIX uses a greedy aggregation approach as follows: Starting with the first data point $x_i$, $i=1$,  in the sorted set of data, we traverse the data points $x_{i+1},x_{i+2},\ldots$ in their order and exploit the inequality
\begin{equation*}
    |\alpha_i - \alpha_j| \leq \|x_i - x_j\|_2 = D(x_i, x_j). 
\end{equation*}
Hence, if $\alpha_j - \alpha_i > \texttt{radius}$ for some index $j > i$, the point $x_j$ and \emph{all the following points} $x_{j+1},x_{j+2},\ldots$ do not need to be considered for the grouping with $x_i$. CLASSIX then moves on to form a new group with the first (smallest index) data point that has not yet been assigned to a group. Each of the computed groups has a special first assigned data point referred to as the \emph{starting point} of the group, which would be $x_i$ in the above example. These points are denoted by $g_k$ and indexed by the group number.

In the merging phase, the starting points (which are already  sorted by their first principal component as they are a subset of the sorted data) are merged into  clusters using the criterion  
\begin{equation*}
    D(g_k, g_\ell) = \| g_k -g_\ell \|_2 \leq \texttt{scale}\times \texttt{radius}.
\end{equation*}
Again, early search termination based on the first principal component can be used for that step. 
The group members of the merged starting points simply become the final clusters. Typically, \texttt{scale} is set to $1.5$ and is not considered a hyperparameter of the algorithm.

Finally, as an optional hyperparameter, CLASSIX provides users with the ability to better handle outliers, or very small clusters, by using the parameter \texttt{minPts}. If $\texttt{minPts} > 1$, all groups in clusters that contain fewer than \texttt{minPts} points will be reassigned to larger nearby clusters. This is performed on the group level, only involving distance computations between group starting points. 

\end{section}

\section{CLASSIX beyond the Euclidean norm}\label{classix:general}

In order to generalize CLASSIX to other distance metrics, it is important to find a suitable search termination criterium to minimize the number of pairwise distance computations. The reverse triangle inequality is a property shared by all norms, including the $p$-norms. 
For any two vectors $u,v\in\mathbb{R}^d$, we have
\begin{equation}\label{eq:revt}
D(u,v) = \| u - v \| \geq \big| \|u\| - \| v \| \big|.
\end{equation}
We now define $\alpha_i = \|x_i\|$ for each data point $x_i$ and, as in the original CLASSIX method, sort these points such that $\alpha_1\leq \alpha_2\leq \cdots \leq \alpha_n$. Then starting with the first data point $x_i$, $i=1$, we traverse the data points $x_{i+1},x_{i+2},\ldots$ in their order as long as $\alpha_i - \alpha_1 \leq \texttt{radius}$. Only these data points need to be considered as candidates for the first group with starting point $x_i$. Among these candidates, data points $x_j$ that satisfy $\|x_j-x_i\|\leq \texttt{radius}$ will be assigned to the first group. Once the first group is completed, we continue finding all points in the neighborhood of the first unassigned data point. This continues until all data points have been assigned to a group. 

A pseudocode for the aggregation phase using an arbitrary norm~$\|\cdot\|$ is given in Algorithm~\ref{classix_m:algorithm}.
\begin{algorithm}
\caption{CLASSIX aggregation in arbitrary norm~$\|\cdot\|$}\label{classix_m:algorithm}
\begin{algorithmic}
    \State $x \gets \texttt{data}$\;
    \State $n \gets \text{number of data points}$\;
    \State $\alpha \gets [\|x_1\|, \|x_2\|, \dots, \|x_n\|]$\;
    \State $\alpha \gets \text{sort}\{\alpha_i\}$\;
    \State $indsort \gets \text{sorting indices of } \alpha$\;
    \State $\text{Sort } x \text{ according to } indsort \text{ along the second axis}$
    \For{$i \gets  1$ to $n$}
        \If{$x_i$ already assigned to a group}
            \State continue
        \EndIf
        \State Start a new group\;
        \State $lastj \gets$ largest index for which $\alpha_{lastj} \leq  \alpha_i +  \texttt{radius}$\;
        \For{$j \gets  i$ to $lastj$}
            \If{$x_j$ already assigned to a group}
                \State continue
            \EndIf
            \If{$\|x_i - x_j\| \leq  \texttt{radius}$}
                \State Add $x_j$ to the current group\;
            \EndIf
        \EndFor
    \EndFor
\end{algorithmic}
\end{algorithm}

The merging phase and \texttt{minPts} filtering remain the same as described in Section~\ref{classix:euclidean}.

It should be noted that the inequality \eqref{eq:revt} can potentially be made sharper by shifting all data points. This is because
\begin{equation}
D(u-c,v-c) = D(u,v) \geq \big| \|u-c\| - \|
v-c\| \big| \label{eq:shift}
\end{equation}
with only the right-hand side depending on $c$. We may therefore try to choose a shift~$c$ so that the right-hand side becomes large on average over all data points. We will explore this in more detail for the Manhattan norm in the following subsection.

\subsection{On the choice of shift for the Manhattan norm}

Let us have a closer look at  \eqref{eq:shift} in the case of the Manhattan ($p=1$) norm. This norm is more robust to outliers than the Euclidean norm and hence particularly important in practice. 

Denoting by $u_i,v_i,c_i$ the components of the vectors $u,v,c$, respectively, we have
\[
\|u - v\|_1 = \sum_{i=1}^d |u_i - v_i| \geq \Big| \sum_{i=1}^d (|u_i-c_i| - |v_i-c_i|)\Big| =: R(c),
\]
and our aim is to find a vector $c$ that maximizes $R(c)$ for this particular choice of $u,v$.
It is straightforward to verify that all the terms in the sum defining $R(c)$ become maximal and positive, and hence maximize $R(c)$, if we choose
\[
\begin{cases}
c_i \leq v_i \quad \text{if} \quad v_i \leq u_i,\\
c_i \geq v_i \quad \text{if} \quad v_i \geq u_i.\\
\end{cases}
\]
Likewise, we could decide to make all terms in the sum most negative, and hence again maximize $R(c)$, by choosing
\[
\begin{cases}
c_i \leq u_i \quad \text{if} \quad u_i \leq v_i,\\
c_i \geq u_i \quad \text{if} \quad u_i \geq v_i.\\
\end{cases}
\]
Put more concisely, we should choose each $c_i$ to be outside the interval $(u_i,v_i)$. But whether to choose a point on the left or right of that interval depends on \emph{all} components~$i$. 

Unfortunately, we do not know in advance what data pairs $u,v$ will be involved in distance computations, and their order will depend on the ``optimal'' shift $c$ we try to find. However, the above two cases suggest that a ``good'' choice for each $c_i$ is to take the component-wise minimum or maximum across all data points. As it appears hard to decide whether to choose the minimum or maximum for each component, we simply recommend choosing $c_i$ as the component-wise minimum of all data points. As this ensures that the shifted data points have only nonnegative entries, we also refer to this as \emph{orthant shifting}.

\paragraph{Example} To illustrate the effect of orthant shifting, consider the two vectors $u = [-2,1]^T$ and $v = [2,-1]^T$. We have $\|u-v\|_1 = 6$ and $| \|u\|_1 - \|v\|_1 | = |3-3| = 0$, and so the reverse triangle inequality merely states the obvious $\| u - v\|_1 \geq 0$. If we choose $c = [ \min\{u_1,v_1\} , \min\{u_2,v_2\}]^T = [-2,-1]^T$, then we get the improved inequality 
\[
\|u - v\|_1 \geq \big| \|u - c\|_1 - \|v - c\|_1 \big| = 2.
\]
Choosing $c = [ \min\{u_1,v_1\} , \max\{u_2,v_2\}]^T = [-2,1]$ gives a perfectly sharp 
\[
\|u - v\|_1 \geq \big| \|u - c\|_1 - \|v - c\|_1 \big| = 6.
\]

We note that there is no guarantee that shifting improves the reverse triangle inequality even when only considering two points $u,v$. Indeed, one can construct examples where shifting worsens the inquality. However, in the large number of  numerical experiments we have performed, we found that orthant shifting usually reduces the number of pairwise distance computations when using an early search termination based on~\eqref{eq:shift}.

\subsection{Efficiency analysis for the Manhattan norm}

Let us now provide some theoretical insights into the effectiveness of the aggregation algorithm in the Manhattan norm. An important factor determining the effectiveness of the aggregation phase, Algorithm~\ref{classix_m:algorithm}, is the \emph{efficiency} of the early search termination bound. This can be phrased as the following question:

\emph{Given a vector $u$ with the Manhattan norm $\alpha_u$, what is the probability that a second vector $v$ with norm $\alpha_v$ is not excluded by the triangle inequality but is at a distance larger than $\texttt{radius}$ from $u$?}

\begin{figure}[ht]
\centering
\resizebox{\columnwidth}{!}{%
\begin{tikzpicture}
  \node[draw, diamond, minimum width=3cm, minimum height=3cm][black, dashed] at (0,0) {};
  \node[draw, diamond, minimum width=7cm, minimum height=7cm] at (0,0) {};
  \node[draw, diamond, minimum width=7cm, minimum height=7cm][black, dotted] at (0,0) {};
  \node[draw, diamond, minimum width=4cm, minimum height=4cm] at (0.5, 1) {};
  \node[draw, diamond, minimum width=4cm, minimum height=4cm, fill=black, opacity=0.1][black, dotted] at (0.5, 1) {};
  \filldraw[black] (0.5, 1) circle (0.03) node[right] {$u$};
  \filldraw[black] (3.5, 0) circle (0.03) node[above right] {($\alpha_u + \texttt{radius}$, 0)};
  \filldraw[black] (0, 3.5) circle (0.03) node[right] {(0, $\alpha_u + \texttt{radius}$)};
  \draw[->, thick] (-4, 0) -- (4, 0) node[anchor=west] {$x$};
  \draw[->, thick] (0, -4) -- (0, 4) node[anchor=south] {$y$};
\end{tikzpicture}%
}
\caption{Comparison of search area vs possible location of vectors satisfying the radius condition.}
\label{fig:1}
\end{figure}
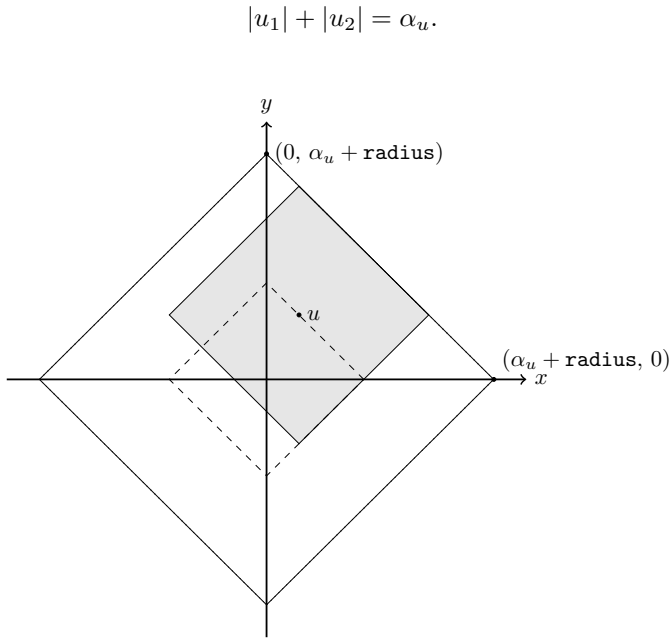

For illustration, we first consider a two-dimensional feature space. 
Assume that we indeed have a vector $u=[u_1,u_2]^T$ with Manhattan norm $\alpha_u$. The possible region in which this vector lies in $\mathbb{R}^2$ is the boundary of the smaller square drawn with dashed lines centered around the origin, as shown in Figure~\ref{fig:1},
\[ 
|u_1| + |u_2| = \alpha_u.
\]

The search termination criterion derived from the reverse triangle inequality restricts the value of $\alpha_v$ to $\alpha_v \leq \alpha_u + \texttt{radius}$, i.e., the candidate points $[x,y]^T$ satisfy $|x| + |y|\leq \alpha_u + \texttt{radius}$, which is the set shown as the larger square centered at the origin. Points $[x,y]^T$ that are at most distance~$\texttt{radius}$ away from~$u$ lie in $|x - u_1| +|y - u_2| \leq \texttt{radius}$, which is indicated by the shaded square centered at the point~$u$. Assuming a uniform distribution of components of the data vectors, in 2D, the search termination efficiency is given by the ratio of the volumes (area) of the shaded and unshaded squares,
\[
\text{efficiency} = \frac{2\texttt{radius}^2}{2(\texttt{radius}+\alpha_u)^2}  = \Big( \frac{\texttt{radius}}{\texttt{radius}+\alpha_u} \Big)^2.
\]


In $d$-dimensional feature space, given a point $u = [u_1,\ldots,u_d]^T$ with score $\alpha_u = \sum_{i=1}^d |u_i|$, the candidate points $v$ must satisfy
\[
\sum_{i = 1}^d |v_i - u_i| \leq \alpha_u + \texttt{radius}.
\]
Hence, all candidate points lie in a $d$-dimensional analogue of the large unshaded polytope shown in Figure~\ref{fig:1}. Similarly, the region containing the actual points $v$ that are a distance at most $\texttt{radius}$ away from $u$ is a $d$-dimensional analogue of the shaded polytope in Figure~\ref{fig:1}.  The two polytopes are  regular (having all internal angles equal) and the ratio between their sides is equal to $(\alpha_u +\texttt{radius})/\texttt{radius}$. Therefore, the ratio of their volumes is given by 
\[
\frac{\texttt{radius}^d}{(\texttt{radius}+ \alpha_u)^d}.
\]

It is clear that the efficiency of an early search termination criterion based on the reversed triangle inequality will degrade as the feature dimension~$d$ increases. It turns out that for certain distances, such as the Tanimoto distance, more efficient criteria can be derived.


\section{CLASSIX in the Tanimoto distance}

As mentioned previously, the Tanimoto distance \eqref{eq:tanimoto} between binary vectors is particularly important in some applications. While the $p=1$ norm distance between binary vectors counts the number of distinct bits in a vector, the Tanimoto distance instead measures the number of common positive bits relative to the joint support of both vectors. As before, we will sort the data points by $\alpha_i = \| x_i \|_1$, which in the context of binary vectors is simply the number of nonzero bits. With these precomputed, we have 
\[
D_{\text{Tanimoto}}(x_i, x_j) = 1 - \frac{x_i \cdot x_j}{\alpha_i + \alpha_j - x_i \cdot x_j} \leq \texttt{radius}.
\]
This gives rise to an efficient way to compute pairwise Tanimoto distances between a query vector $x_i$ and many data points, say, $Y = [x_j, x_{j+1}, \ldots, x_n]^T$,  in form of a single matrix-vector multiply $Y\cdot x_i$.

There are two key changes that we propose over the CLASSIX algorithm in the $p=1$ norm. First, when working with bit vectors and the Tanimoto distance, orthant shifting of the data points is not necessary, so we can skip this step. Secondly, and more importantly, we can utilize an improved search termination criterion as discussed below.

\begin{subsection}{A sharper search bound for the Tanimoto distance}\label{tanimoto:bound}

For Morgan fingerprints in chemical databases, and the Tanimoto distance metric, a search termination criterion that improves upon the reverse triangle inequality was proposed by Baldi et al.~\cite{swamidass2007bounds, baldi2009intersection}.

Let us again define the score of a point $\alpha_i = \|x_i\|_1$ and assume that all points are sorted by their score. For a pair of points $x_i$ and $x_j$ with scores $\alpha_i$ and $\alpha_j$, respectively, the minimal Tanimoto distance possible is 
$$
(D_{\text{Tanimoto}})_{\min} = 1 - \frac{\min(\alpha_i, \alpha_j)}{\max(\alpha_i, \alpha_j)}.
$$ 
This distance is attained if all the $1$-bits in the two vectors overlap. 

Consider a query vector $x_i$ with score  $\alpha_i$, and bisect the search space into cases where $\alpha_j < \alpha_i$ and $\alpha_j > \alpha_i$. Since our points are sorted in order of their scores, and we start aggregation from the point with the lowest score, as explained in Section~\ref{classix:euclidean}, we only need to consider the region where $\alpha_j > \alpha_i$, which gives the condition
\begin{equation}
D_{\text{Tanimoto}}(x_i,x_j) \geq  1 - \frac{\alpha_i}{\alpha_j}.   \label{eq:tanineq}
\end{equation}
Hence, we can reject all points~$x_j$ whose scores do not meet the criterion 
\[
\alpha_j \leq \frac{\alpha_i}{1-\texttt{radius}}.
\]
This criterion turns out to be sharper than the reverse triangle inequality, as we show in the next subsection.

\end{subsection}

\begin{subsection}{Baldi inequality is sharper  than  triangle inequality}\label{tanimoto:baldi_proof}

In the following, it will be more convenient to consider the Tanimoto similarity between binary vectors $u$ and $v$, rather than distance, defined by \begin{align*}
\mathrm{sim}(u, v) & := 1 -D_\mathrm{Tanimoto} = \frac{\|u \land v\|}{\|u\lor v\|} = \frac{u \cdot v}{\|u\| +\|v\|- u \cdot v},
\end{align*}
where in this subsection $\| \cdot \|$ denotes the $1$-norm (which is the same as the count of nonzero bits). We also assume throughout that $u \lor v \neq 0$. It is known that the Tanimoto distance is a proper distance metric that satisfies the triangle inequality~\cite{Lipkus1999}.  Hence, we can lower-bound the Tanimoto distance between $u$ and $v$ via  a third vector $w\neq 0$ using the reverse triangle inequality,
\begin{align*}
    D_\mathrm{Tanimoto}(u, v)   &\geq |D_\mathrm{Tanimoto}(u, w) - D_\mathrm{Tanimoto}(v, w)|,
\end{align*}
or equivalently, in terms of Tanimoto similarity
\begin{align*}
    \mathrm{sim}(u, v)   &\leq 1 - |\mathrm{sim}(v, w) - \mathrm{sim}(u, w)|.
\end{align*}

We define the (set) difference for binary vectors as
\[
u\setminus v := u - (u \land v).
\]
For three binary vectors $u,v,w$ the following relation between their intersections can easily be verified:
\begin{equation}\label{eq:bineq}
    u \land v = (w\land u\land v)\lor ((u \setminus w)\land (v \setminus w)).
\end{equation}
Let $\beta = u\land w $ and $\gamma = v \land w$, then
\begin{align*}
    \|w\land u\land v\| &\leq \min(\|u\land w\|, \|v\land w\|) \\
    &= \min(\|\beta\|, \|\gamma\| ),
\end{align*}
and 
\begin{align*}
    \|(u \setminus w)\land (v \setminus w)\| &\leq \min(\|u \setminus w\|, \|v \setminus w\|) \\
    &= \min(\|u - \beta\|, \|v -  \gamma\|).
\end{align*}
Using \eqref{eq:bineq} and the fact that $\|u \lor v\| \leq \|u\| + \|v\|$, we obtain 
\begin{align*}
    \|u\land v\|    &\leq \|w\land u\land v\|+\|(u \setminus w)\land (v \setminus w)\|\\
                    &\leq \min(\|\beta\|, \|\gamma\|) + \min(\|u - \beta\|, \|v - \gamma\|).
\end{align*}
This leads to the bound on the Tanimoto similarity between $u$ and $v$ as
\begin{equation}
\begin{aligned}[t]
    &\mathrm{sim}(u, v) \\
    &\quad \leq \frac{\min(\|\beta\|, \|\gamma\|) + \min(\|u - \beta\|, \|v - \gamma\|)}{\|u\|+\|v\|- \min(\|\beta\|, \|\gamma\|) - \min(\|u - \beta\|, \|v - \gamma\|)}.
\end{aligned}
\label{eq:baldiineq2}
\end{equation}  
This is the general inequality introduced by Baldi et al.~\cite{baldi2009intersection}, and it reduces to the inequality \eqref{eq:tanineq} when $w = [1,1,\ldots,1]^T$: in this case $\|\beta\|=\|u\|=\alpha_u$ and $\|\gamma\|=\|v\|=\alpha_v$, while $\|u-\beta\|=\|v-\gamma\|=0$.

We now show that the expression on the right of \eqref{eq:baldiineq2} does not exceed the bound from the triangle inquality; that is,
\begin{equation*}
\begin{aligned}
&\frac{\min(\|\beta\|, \|\gamma\|) + \min(\|u - \beta\|, \|v - \gamma\|)}
      {\|u\|+\|v\| - \min(\|\beta\|, \|\gamma\|) - \min(\|u - \beta\|, \|v - \gamma\|)} \\
&\quad \quad \leq 1 - |\mathrm{sim}(v, w) - \mathrm{sim}(u, w)|
\end{aligned}
\end{equation*}
for any choice of $w$. 

The LHS and RHS are symmetric in $u,\beta$ and $v,\gamma$ (i.e., pairs~$(u,\beta)$ and $(v,\gamma)$ can be interchanged without changing the expressions). Without loss of generality, we can assume that $\|w\lor u\| = k\|w\lor v\|$, where $k \geq 1$. This leads to 
\[
    \|w\|+\|u\| - \|\beta\| = k(\|w\|+\|v\| - \|\gamma\|),
\]  
or equivalently, 
\begin{align*}
    \|u\| - \|\beta\| &= (k-1)\|w\| + k\|v\| -k\|\gamma\|\\
    &= (k-1)(\|w\| + \|v\| -\|\gamma\|) + \|v\| - \|\gamma\|.
\end{align*}
This implies $\|u - \beta\| = \|u\| - \| \beta \| \geq \|v\| - \|\gamma\|$. 
We can then rewrite the inequality we are trying to prove equivalently as
\begin{align*}
    &\frac{\min(\|\beta\|, \|\gamma\|) +  \|v\| - \|\gamma\|}{\|u\| + \|v\| - \min(\|\beta\|, \|\gamma\|) - \|v\| + \|\gamma\|} \\
    &\quad \quad\leq 1 - \big| \frac{\|\beta\|}{\|w \lor u\|} - \frac{\|\gamma\|}{\|w\lor v\|} \big|\\
    &\quad \quad = 1 -\big| \frac{\|\beta\|}{\|w \lor u\|} - \frac{k\|\gamma\|}{\|w\lor u\|} \big|,
\end{align*}
or, 
\begin{align*}
    \frac{\|v\| + \min(\|\beta\|, \|\gamma\|)-\|\gamma\|}{\|u\| - \min(\|\beta\|, \|\gamma\|) + \|\gamma\|} 
    &\leq \frac{\|w \lor u\| - |\|\beta\| - k\|\gamma\||}{\|w \lor u\|}.
\end{align*}
Since the denominators are strictly positive, we can cross-multiply them to obtain \:
\begin{align*}
    &(\|w\lor u\|)(\|v\|+\min(\|\beta\|, \|\gamma\|)-\|\gamma\|)\\ 
    &\quad \quad \leq (\|w\lor u\| - |\|\beta\| -k\|\gamma\||)(\|u\| - \min(\|\beta\|, \|\gamma\|)+\|\gamma\|),
\end{align*}
or equivalently,
\begin{align*}
    &(\|w\lor u\|)(\|v\| - \|u\| +2\min(\|\beta\|, \|\gamma\|) - 2\|\gamma\|) \\
    &\quad \quad \leq - |\|\beta\| - k\|\gamma\||(\|u\| - \min(\|\beta\|, \|\gamma\|) + \|\gamma\|).
\end{align*}
Multiplying both sides by $-1$, we get \:
\begin{align*}
    &|\|\beta\| - k\|\gamma\||(\|u\| - \min(\|\beta\|, \|\gamma\|) + \|\gamma\|)\\
    &\quad \quad \leq (\|w\lor u\|)(\|u\| - \|v\| - 2\min(\|\beta\|, \|\gamma\|) + 2\|\gamma\|).
\end{align*}
For our algorithm, we choose $w = (1, 1, 1, \dots)$, which results in the special case $k=1$, since $u\lor w = v \lor w = w$. If $k=1$, then $\|u\| - \|\beta\| = \|v\| - \|\gamma\|$. We have the following two subcases to examine and $\|\beta\| = \alpha_u$ and $\|\gamma\| = \alpha_v$, the scores of the vectors $u$ and $v$:
\begin{itemize}
    \item In the symmetric case, $\alpha_u = \alpha_v$ implies $\|u\| = \|v\|$. In this case, both sides are equal. 
    \item In the non-symmetric case, without any loss of generality, we can assume $\alpha_v<\alpha_u$. Then, the inequality holds as
    \begin{align*}
        & |\|\beta\| - \|\gamma\||(\|u\| - \|\gamma\| + \|\gamma\|) \\
        \quad &\leq (\|w\lor u\|)(\|u\| - \|v\| - 2 \|\gamma\| + 2\|\gamma\|),
    \end{align*}
    or, 
    \begin{align*}
        |\|\beta\| - \|\gamma\||\cdot \|u\| &\leq (\|w \lor u\|)(\|u\| - \|v\|), \\
        \|u\|&\leq \|w \lor u\|,
    \end{align*}
    which is always true, since $\|w\lor u\|=\|u\| +\|w\| - \|u\cdot w\|$, and $\|u\| \leq \|u\| +\|w\| -\|u\cdot w\|$.
\end{itemize}



\end{subsection}

\subsection{Probabilistic analysis}\label{tanimoto:analysis}

We now perform an ``average-case'' analysis of the efficiency if the Baldi search condition~\eqref{eq:tanineq}. 
Let us assume that $x_i\in\{0,1\}^d$ is fixed and that a large number of data points $x_j$ are generated by flipping each bit of $x_i$ at random with probability $p \ll 1$. We can then consider the following two probabilities: 
$$P_1 = \mathbb{P}(\alpha_i s \leq \alpha_j \leq \alpha_i /s ),$$
and 
$$P_2 = \mathbb{P}(\mathrm{sim}(x_i,x_j) \geq s).$$
Here, $\mathrm{sim}(\cdot)$ refers to Tanimoto similarity. We clearly have $P_1 \geq P_2$, and the quotient $P=P_2/P_1$ can be interpreted as a conditional probability: 

\emph{What is the probability that two data points $x_i, x_j$ have a Tanimoto similarity of at least $s$ given that $\alpha_i s \leq \alpha_j \leq \alpha_i /s$?}

Ideally, we want $P=1$ because this would mean that the early search termination criterion perfectly identifies all data points within the search radius and only those.

It is straightforward to derive the probability distribution function $\mathbb{P}(\alpha_j = k)$. The vector $x_i$ has exactly $\alpha_i = \|x_i\|_1$ non-zero components (all ones) and $d-\alpha_i$ zeros. 
For $x_j$ to have exactly $k$ nonzeros, we can start with $x_i$ and flip $\ell = \max\{0, \alpha_i - k\},\ldots,\alpha_i$ of its ones to zeros, and then $k+\ell-\alpha_i$ of its zeros to ones. Denoting by $f(k,n,p) =  {n \choose k} p^k (1-p)^{n-k}$ the probability mass function of getting exactly $k$ successes in $n$ independent Bernoulli trials with the same rate $p$, we therefore have
\[
\mathbb{P}(\alpha_j = k) = 
\sum_{\ell = \max\{0, \alpha_i - k\}}^{\alpha_i}
f(\ell,\alpha_i,p) \cdot f(k+\ell-\alpha_i, d-\alpha_i,p).
\]
Summing over all possible values of $\alpha_j$, that is, $\alpha_i s\leq \alpha_j \leq \alpha_i/s$, we get
\begin{align*}
    P_1 &=\sum_{k = \alpha_i s}^{\alpha_i / s} \mathbb{P}(\alpha_j=k) \\
        &=  \sum_{k = \alpha_i s}^{\alpha_i / s}\ 
\sum_{\ell = \max\{0, \alpha_i - k\}}^{\alpha_i}
f(\ell,\alpha_i,p) \cdot f(k+\ell-\alpha_i, d-\alpha_i,p).
\end{align*}

For computing $P_2$, we sum over all possible values of $\ell$, such that it satisfies the constraint $\mathrm{sim}(x_i, x_j)\geq s$ by observing the following relation.

If a given vector $x_j$ has exactly $\alpha_j = k$ nonzeros, and it was generated from~$x_i$ (which has $\alpha_i$ nonzeros) by flipping $\ell$ of its ones to zeros and then $k + \ell - \alpha_i$ of its zeroes to ones, we have $x_i\cdot x_j = \alpha_i - \ell$. Using the dot product formula for Tanimoto similarity
\[
\mathrm{sim}(x_i, x_j) = \frac{x_i\cdot x_j}{\|x_i\| + \|x_j\| - x_i \cdot x_j},
\]
the similarity score between $x_i$ and $x_j$ will be equal to
\begin{align*}
    \mathrm{sim}(x_i, x_j) &= \frac{\alpha_i - \ell}{\alpha_i + \alpha_j - (\alpha_i -\ell)} = \frac{\alpha_i - \ell}{\alpha_j + \ell}\,.
\end{align*}
Now, we can constrain the values of $\ell$ using the criterion $\mathrm{sim}(x_i, x_j) \geq s$, as 
\begin{align*}
    \frac{\alpha_i - \ell}{\alpha_j + \ell}  \geq s \ \Longleftrightarrow  
    \ell  \leq \frac{\alpha_i - \alpha_j s}{1+s}.
\end{align*}
So, the probability that $\ell$ bits are flipped, and that simultaneously the constraint $\mathrm{sim}(s_i, s_j)\geq s$ is satisfied, is
\[
\begin{aligned}
\mathbb{P}(\alpha_j=k \mid \mathrm{sim}(x_i, x_j) \geq s) 
&= \sum_{\ell = \max\{0, \alpha_i - k\}}^{\left[\frac{\alpha_i - s k }{1+s}\right]} 
f(\ell,\alpha_i,p) \\
&\quad \cdot f(k+\ell-\alpha_i, d-\alpha_i,p).
\end{aligned}
\]

We should note that the lower limit of the sum will never be greater than the upper limit since the values of $k$ are bound by $\alpha_i s \leq k \leq \alpha_i/s$. Then, $P_2$ simply becomes
\begin{align*}
    P_2 &=\sum_{k = \alpha_i s}^{\alpha_i / s} \mathbb{P}(\alpha_j=k \mid \mathrm{sim}(x_i, x_j)\geq s) \\
        &=  \sum_{k = \alpha_i s}^{\alpha_i / s}\ 
\sum_{\ell = \max\{0, \alpha_i - k\}}^{[\frac{\alpha_i - s k }{1+s}]}
f(\ell,\alpha_i,p) \cdot f(k+\ell-\alpha_i, d-\alpha_i,p),
\end{align*}
a subsum of the expression we derived for $P_1$ (and hence, clearly, $P_2\leq P_1$).

\subsection{Performance evaluation of the search termination}

\begin{figure}[htbp]
    \centering
    \includegraphics[width=\columnwidth]{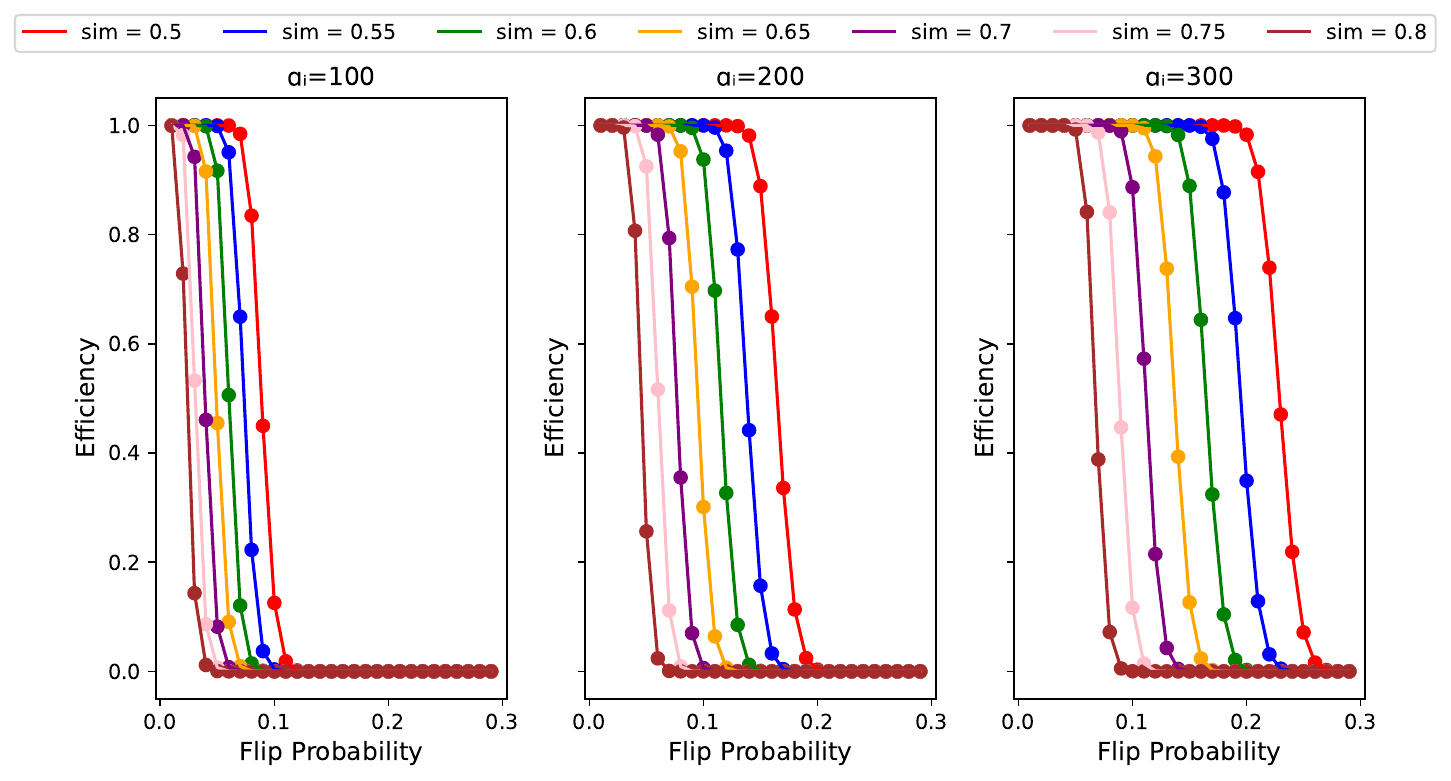}
    \caption{Efficiency of the search termination criterion based on different flip probabilities and similarity thresholds for different values of $\alpha_i$, as a result of direct computation of probabilities.}
    \label{fig:efficiency_computation}
\end{figure}

We can now evaluate the efficiency $P = P_2/P_1$ of the proposed search termination criterion through numerical computation and simulation. First, using the expressions derived above, we calculate the efficiency for varying flip probabilities $p$, the seed vector scores~$\alpha_i$, and  similarity parameters~$s$, while fixing the dimensionality of the data at~$d = 1,000$. The results are presented in Figure~\ref{fig:efficiency_computation}.

\begin{figure}[htbp]
    \centering
    \includegraphics[width=\columnwidth]{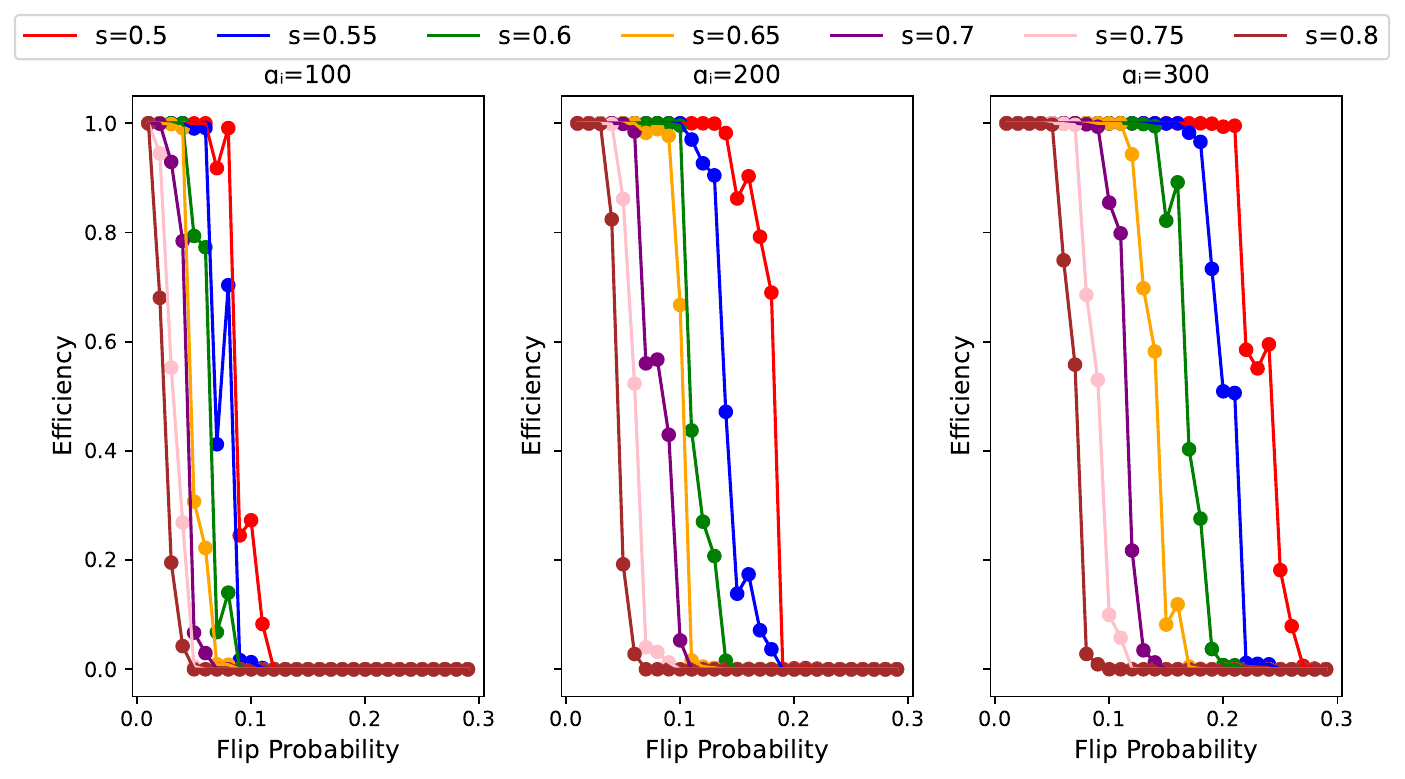}
    \caption{Efficiency of the search termination criterion based on different flip probabilities and similarity thresholds for different values of $\alpha_i$, as a result of simulations on synthetic data.}
    \label{fig:efficiency_simulation}
\end{figure}

Second, we perform a simulation-based estimation of efficiency by generating a single cluster consisting of $n = 10,000$ data points for various values of $p$ and $\alpha_i$, again with the data dimension fixed at $d = 1,000$. For each setting, we compute the ratio of data points filtered by the search termination criterion to the total number of points lying within a Tanimoto distance of $\texttt{radius} = 1 - s$ from the seed vector $x_i$, where $s$ is varied. The results are illustrated in Fig.~\ref{fig:efficiency_simulation}. Details regarding the data generation process can be found in Section~\ref{experiments}, Algorithm~\ref{experiments:dataalgorithm}. (The code used to generate the aforementioned figure can be found on the GitHub repo for this paper.)

The flip probability $p$ characterizes the dispersion within a data cluster, where a higher value of $p$ corresponds to increased cluster spread. Conversely, the similarity parameter~$s$ quantifies the compactness of the desired clusters, with larger values indicating tighter groupings (i.e., smaller clustering radius). Fig.~\ref{fig:efficiency_computation} and Fig.~\ref{fig:efficiency_simulation} indicate that the search termination criterion improves when applied to clusters with lower dispersion and under more relaxed similarity constraints. Furthermore, we observe a consistent increase in efficiency when the seed vector score~ $\alpha_i$ increases. This trend is evidenced by the rightward shift in the efficiency curves in both Fig.~\ref{fig:efficiency_computation} and Fig.~\ref{fig:efficiency_simulation} as $\alpha_i$ increases. The search termination criterion becomes more efficient for increasing $\alpha_i$ up to $\alpha_i = d/2$. After that point, the criterion is redundant for certain values of $s$, since we end up not filtering any data points at all, i.e.,~almost all data points lie within a Tanimoto distance of $1-s$ to $x_i$.

\newpage

\begin{section}{Experiments}\label{experiments}

\sloppy
In this section, we demonstrate the scalability of CLASSIX Tanimoto (referred to as CLASSIX\_T) with respect to data size, data dimensionality, number of clusters, data separability by pop-count, and sensitivity to the \texttt{radius} parameter. We generate synthetic data by first generating a number (\texttt{num\_clusters})   of $d$-dimensional binary vectors. We call these vectors the \emph{seeds} for the clusters. Then, for each seed vector $x_i$, we generate new samples in the cluster by flipping its elements with a probability $p=0.1 \times \frac{\|x_i\|_1}{d}$. We repeat this process $k$ times for each seed vector and generate a total of $k\times \texttt{num\_clusters}$ data points. The complete algorithm used to generate the test data is given in Algorithm~\ref{experiments:dataalgorithm}. An implementation of the algorithm can be found on our repository, as the function \texttt{generate\_data}.

\begin{algorithm}
\caption{Data generation algorithm}\label{experiments:dataalgorithm}
\begin{algorithmic}
    \State $data \gets \texttt{empty list}$\;
    \For{$i \gets  1$ to $\texttt{num\_clusters}$}
        \State $x_i \gets \text{generate random boolean vector}$\;
        \State add seed vector $x_i$ to the data\;
        \For{$j \gets 1$ to $k$}
            \State $flip\_mask \gets \text{mask that represents flipped bits}$\;
            \State $y = \mathrm{XOR}(x_i, flip\_mask)$\;
            \State add point $y$ to the data\;
        \EndFor
    \EndFor
\end{algorithmic}
\end{algorithm}

\begin{subsection}{Dependence on data size}\label{tanimoto:data_size}

We compare the performance of CLASSIX\_T with the \texttt{scikit-learn} Python implementations of DBSCAN and OPTICS on synthetic data generated using Algorithm~\ref{experiments:dataalgorithm}. We generate $k$ data points in each cluster, keeping the number of clusters constant at $10$, varying $k$ from $1,000$ to $10,000$. The dimension~$d$ is fixed at $d=1,000$. The hyperparameters for the tested methods are set as follows:

\begin{center}
\begin{tabular}{l l} 
\toprule
 Algorithm & Hyperparameters \\
 \midrule
 CLASSIX & $\texttt{radius} = 0.4, \ \texttt{minPts} = 5$ \\ 
 
 DBSCAN & $\texttt{eps} = 0.4, \ \texttt{min\_samples} = 5$ \\
 
 OPTICS & $\texttt{max\_eps} = 0.4, \ \texttt{min\_cluster\_size} = 5$ \\
\bottomrule
\end{tabular}
\captionof{table}{Hyperparameters for the comparisons}
\label{table:hyperparameters}
\end{center}

The clusters computed by all methods are perfect when measured in the Adjusted Rand Index (ARI). ARI values lie within $[-1, 1]$ and scores greater than $0$ indicate meaningful clusters. The clustering runtimes of DBSCAN and OPTICS are very large compared to CLASSIX\_T, differing by roughly two orders of magnitude; see Fig.~\ref{fig:2}. To give an example, for $n=100,000$ data points, CLASSIX\_T takes $28\,s$, compared to $6,321\,s$ and $17,508\,s$ for DBSCAN and OPTICS, respectively. Hereafter, we limit our performance comparison on synthetic data to CLASSIX\_T and DBSCAN only, given that OPTICS appears to consistently perform worse than the others.

\begin{figure}[htbp]
  \centering
  \includegraphics[width=\columnwidth]{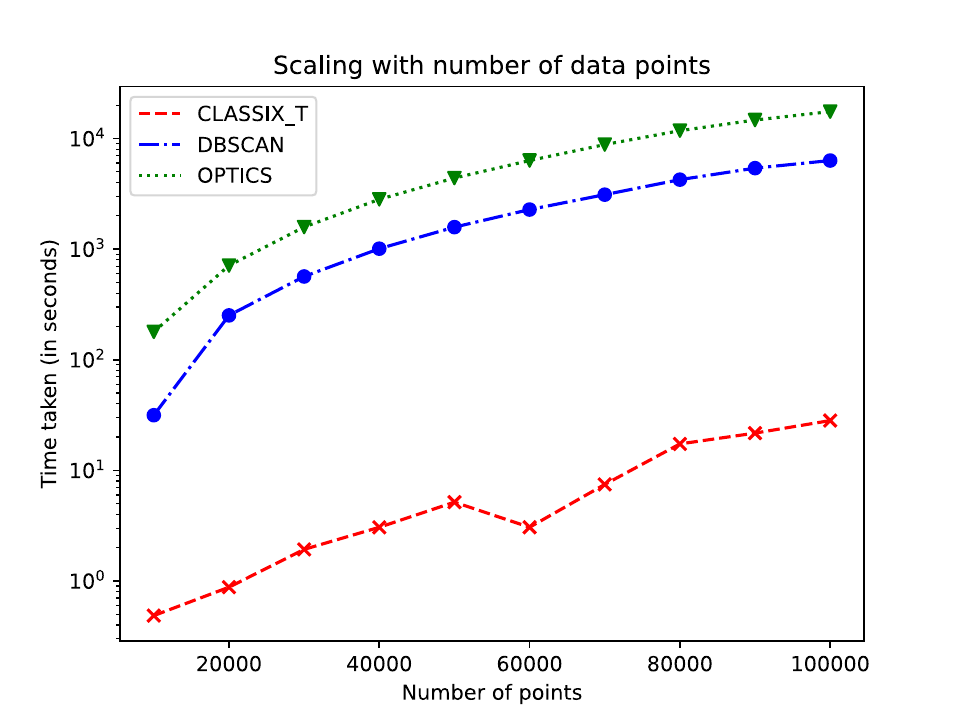}
  \caption{Runtime of CLASSIX\_T (red crosses), DBSCAN (blue dots) and OPTICS (green triangles) as the number of data points, $n$, varies. The data dimension $d=1,000$ and number of clusters $\texttt{num\_clusters} = 10$ are kept fixed.}
  \label{fig:2}
\end{figure}

\end{subsection}

\begin{subsection}{Dependence on the data separability}\label{tanimoto:separability}

We generate data keeping the dimension constant at $d=2,000$. The number of clusters is constant at $10$, and each cluster contains $1,000$ points. We vary the range of seed vector pop-count, which affects the efficiency of CLASSIX\_T's search termination criterion (see Section~\ref{tanimoto:analysis}). The score of the first seed vector is $\alpha_{min} = 50$, and the others are generated so that each seed vector has a score $\beta$ higher than the previous, with the last seed vector score $\alpha_{max} = 9\times\beta + \alpha_{min}$. The clustering runtimes of CLASSIX\_T and DBSCAN for $\beta = 0, 1, \dots, 9$ are shown in Fig.~\ref{fig:3}. We expect a shorter runtime for the data in which seed vectors are more separated by their scores, i.e., for larger values of $\beta$. The hyperparamters used for CLASSIX\_T are $\texttt{radius} = 0.35$ and $\texttt{minPts}=5$, and for DBSCAN $\texttt{eps}=0.35$ and $\texttt{min\_samples} = 5$.

\begin{figure}[htbp]
  \centering
  \includegraphics[width=\columnwidth]{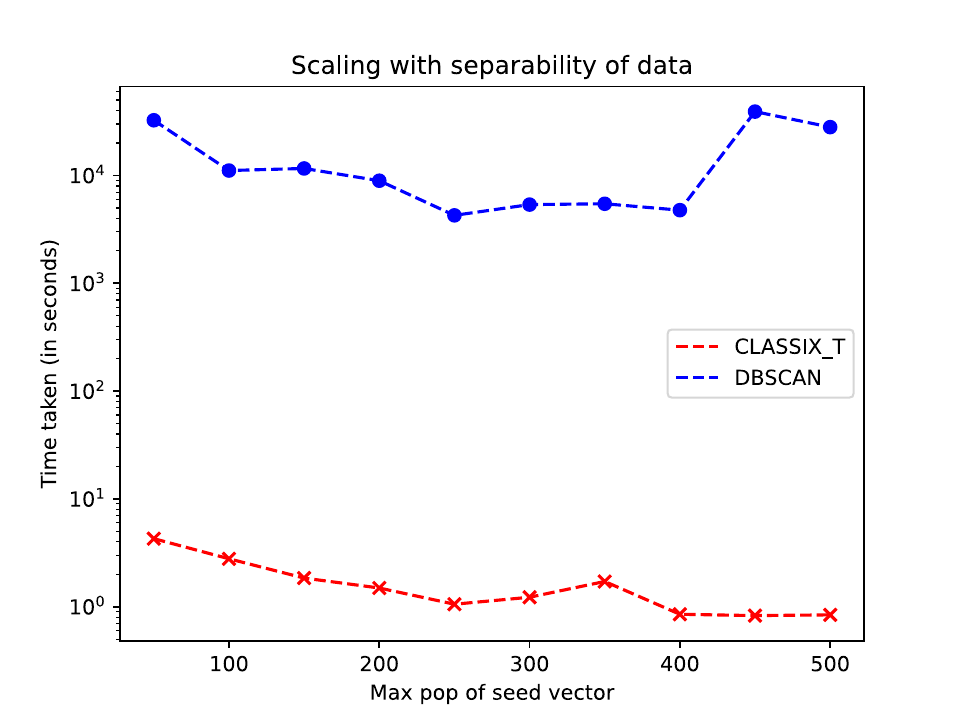}
  \caption{Scaling of CLASSIX\_T (red crosses) and DBSCAN (blue dots) with the separability of the data with respect to the pop-count scores. Total number of data points is kept constant $n=50,000$.}
    \label{fig:3}
\end{figure}

We observe that the runtime decreases as the clusters becomes more separated by their score, up to a certain point after which we do not observe any further decrease in runtimes. In order to explain this, recall our observations from Fig~\ref{fig:efficiency_computation} in section~\ref{tanimoto:analysis}. We observed that the efficiency of the search termination algorithm is higher for larger values of $\alpha_i$. As we allow for vectors with larger $\alpha$~scores, the average efficiency increases, decreasing the runtime. Once we reach a point where most of the vectors have very large $\alpha$~values, they all fall under the ``highly efficient section'' of the plot. Further increasing the difference in scores of the seed vectors does not yield further speed-up. Instead, only small fluctuations due to randomness of the data, especially the number of groups created during the aggregation phase, will determine the relative performance.
    
\end{subsection}

\begin{subsection}{Dependence on data dimension}\label{tanimoto:data_dimension}

We now generate a dataset with fixed numbers of clusters and data points, with $\texttt{num\_clusters} = 10$ and $n=10,000$, respectively, and vary the data dimension $d$ from  $500$ to $5,000$ in steps of $500$. We run CLASSIX and DBSCAN on the generated data and compare their performance. The hyperparameters used can be found in Table \ref{table:hyperparameters}. The runtimes of both CLASSIX\_T and DBSCAN scale approximately linearly with the data dimension, as shown in Fig.~\ref{fig:4}. However, CLASSIX\_T has a much smaller slope compared to DBSCAN.

\begin{figure}[htbp]
  \centering
  \includegraphics[width=\columnwidth]{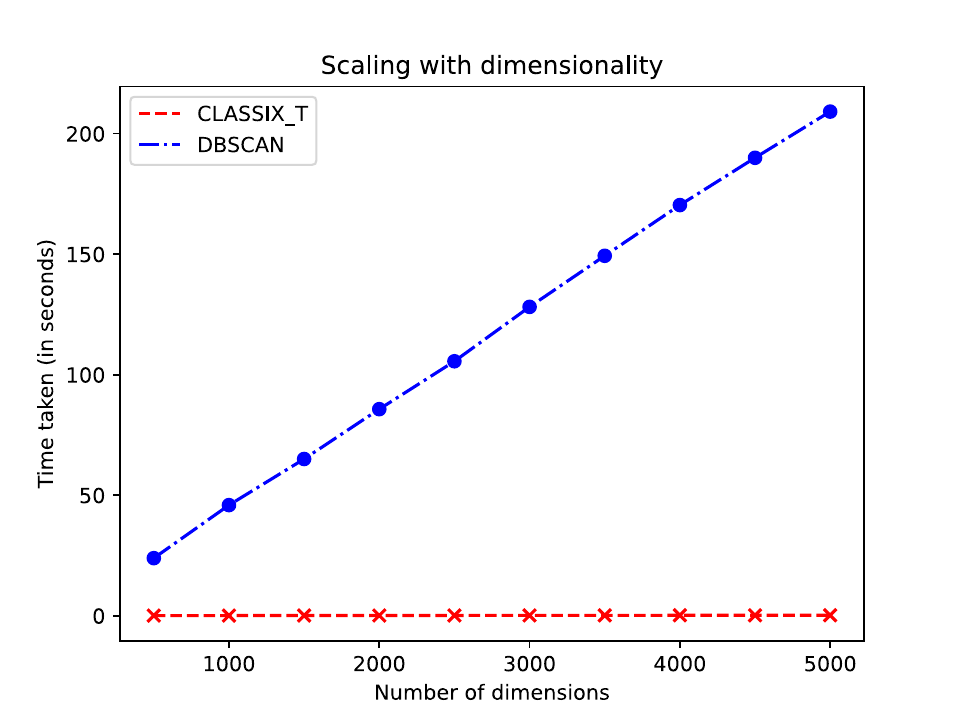}
  \caption{Runtime of CLASSIX\_T (red) and DBSCAN (blue) with respect to the dimension~$d$, keeping the number $n=10000$ of data points fixed.}
  \label{fig:4}
\end{figure}

\end{subsection}

\begin{subsection}{A real-world chemical fingerprint dataset}
We used real chemical fingerprint data compiled by Greg Landrum, published at \cite{greglandrumrdkitblog2019}, to test the performance of CLASSIX\_T and compare it with the Taylor--Butina clustering method \cite{taylor-butina-clustering-1999} available in the RDKit framework \cite{rdkit-framework} at \url{https://www.rdkit.org/}, and DBSCAN. The dataset contains $91,662$ molecules divided into $34$ clusters. The Taylor--Butina method requires  $1856.28s$ to cluster this dataset, DBSCAN takes  $5064.84s$, and CLASSIX\_T takes $61.83s$. The fingerprints were generated using the Morgan fingerprint generator from RDKit with parameters $\texttt{r}=2$ and $\texttt{nBits}=1024$. The hyperparameters were determined using a grid search strategy: the Taylor--Butina parameter $\texttt{cutoff} \in [0.2, 0.5]$ in steps of $0.025$; DBSCAN $\texttt{eps} \in [0.2, 0.5]$ in steps of $0.025$ and $\texttt{min\_samples} \in [10, 100]$ in steps of $5$; CLASSIX\_T $\texttt{radius} \in (0.2, 0.5)$ in steps of $0.025$ and $\texttt{minPts} \in [10, 100]$ in steps of $5$. The best found parameters and the highest achieved ARI scores are given in Table~\ref{tab:greglandrumdata}.

\begin{table}[t]
\centering
\resizebox{\columnwidth}{!}{%
\begin{tabular}{l l l l}
\toprule
Algorithm & Hyperparameters & Runtime & ARI \\
\midrule
Taylor--Butina & $\texttt{cutoff} = 0.35$ & $1856.28\,\text{s}$ & $0.017$ \\
DBSCAN & $\texttt{eps} = 0.35,\ \texttt{min\_samples} = 50$ & $5064.84\,\text{s}$ & $0.089$ \\
CLASSIX\_T & $\texttt{radius} = 0.4,\ \texttt{minPts} = 50$ & $61.83\,\text{s}$ & $0.130$ \\
\bottomrule
\end{tabular}%
}
\caption{Comparison of Taylor--Butina, CLASSIX\_T and DBSCAN on Greg Landrum's molecule dataset.}
\label{tab:greglandrumdata}

\end{table}

CLASSIX\_T is the fastest algorithm and beats DBSCAN by a factor of 80 while generating a better clustering in terms of ARI. Taylor--Butina clustering is faster than DBSCAN, but produces clusters of much lower quality. It is also worth noting that both DBSCAN and the RDKit implementation of Taylor--Butina use bitwise operations for computing the Tanimoto distance, while we use integer arithmetic. CLASSIX\_T could certainly be further optimised by incorporating bitwise operations for computing the Tanimoto distance. 

\end{subsection}

\begin{subsection}{Experiments with CLASSIX in the Manhattan norm}

We next discuss the scalability of CLASSIX\_M algorithm by comparing it to available Manhattan metric implementations of DBSCAN and OPTICS in \texttt{scikit-learn}, using the popular IRIS and Banknote datasets. 
The results are presented in Table~\ref{tab:fulldimensiondata}. The best found hyperparameters and the code to obtain these can be found in our GitHub.

CLASSIX\_M achieves the best ARI score out of the three benchmarked algorithms. The hyperparameters were discovered using a grid search in two dimensions for each algorithm. 
For CLASSIX\_M our grid search entailed $\texttt{radius} \in (0.005, 0.1)$ in steps of $0.005$, $\texttt{minPts} \in (5, 50)$ in steps of $5$ in the IRIS data; $\texttt{radius} \in (0.005, 0.1)$ in steps of $0.005$, $\texttt{minPts} \in (5, 100)$ in steps of $5$ in the Banknote data.

For DBSCAN $\texttt{eps} \in (0.25, 5)$ in steps of $0.25$, $\texttt{min\_samples} \in (5, 50)$ in steps of $5$ in the IRIS data; and $\texttt{eps} \in (0.05, 1.05)$ in steps of $0.05$, $\texttt{min\_samples} \in (5, 50)$ in steps of $5$ in the Banknote data.

For OPTICS $\texttt{max\_eps} \in (0.1, 2)$ in steps of $0.1$, $\texttt{min\_samples} \in (5, 50)$ in steps of $5$ in the IRIS data; and $\texttt{max\_eps} \in (1, 2.5)$ in steps of $0.05$, $\texttt{min\_samples} \in (5, 50)$ in steps of $5$ in the Banknote data.

\begin{table*}[t]
\centering

\begin{tabular}{l c c c c c c} 
\toprule
Dataset 
& \multicolumn{2}{c}{DBSCAN} 
& \multicolumn{2}{c}{OPTICS} 
& \multicolumn{2}{c}{CLASSIX\_M} \\
\cmidrule(lr){2-3} \cmidrule(lr){4-5} \cmidrule(lr){6-7}
& Best ARI & Runtime 
& Best ARI & Runtime 
& Best ARI & Runtime \\
\midrule
IRIS & 0.57 & \textbf{0.003} & 0.56 & 0.030 & \textbf{0.62} & 0.025 \\
Banknote & 0.84 & \textbf{0.023} & 0.045 & 0.36 & \textbf{0.90} & 0.052 \\
\bottomrule
\end{tabular}
\caption{Comparison of best achieved ARI and their corresponding runtimes of DBSCAN, OPTICS and CLASSIX\_M, all using the Manhattan norm.}
\label{tab:fulldimensiondata}
\end{table*}

While all runtimes for these small datasets are in the order of milliseconds, it must be noted that  CLASSIX\_M takes longer than DBSCAN. These datasets produce a large number of outliers that need to be handled. These take up most of the computation time. Also, The CLASSIX\_M algorithm does not use sparse vector products, unlike CLASSIX\_T, which leads to increased time consumption.

\end{subsection}

\begin{subsection}{Experiments with CLASSIX Manhattan and UMAP}

For the MNIST (Fig.~\ref{fig:7}) dataset, we use  UMAP \cite{mcinnes2020umapuniformmanifoldapproximation} as a dimension reduction tool to perform the clustering in a reduced 2-dimensional feature space. We used the UMAP parameters $ \texttt{n\_components}=2, \texttt{metric}=\text{manhattan}$. Other parameters were left at their default settings. We have included the performance of CLASSIX\_M in the IRIS (Fig.~\ref{fig:5}) and Banknote (Fig.~\ref{fig:6}) datasets using dimension reduction for completeness. 

The hyperparameters were optimized for maximum ARI by performing a grid search in two dimensions for each algorithm. The hyperparameter ranges for  CLASSIX\_M were $\texttt{radius} \in (0.1, 0.5)$ in steps of $0.025$, $\texttt{minPts} \in (0, 50)$ in steps of $5$ for IRIS; $\texttt{radius} \in (0.1, 0.5)$ in steps of $0.025$, $\texttt{minPts} \in (0, 50)$ in steps of $5$ for Banknote; and $\texttt{radius} \in (0.01, 0.1)$ in steps of $0.005$, $\texttt{minPts} \in (0, 50)$ in steps of $5$ for the MNIST dataset. 

For DBSCAN $\texttt{eps} \in (0.1, 0.5)$ in steps of $0.025$, $\texttt{min\_samples} \in (0, 50)$ in steps of $5$ for IRIS; $\texttt{eps} \in (0.1, 0.5)$ in steps of $0.025$, $\texttt{min\_samples} \in (0, 50)$ in steps of $5$ for Banknote; and $\texttt{eps} \in (0.1, 1)$ in steps of $0.025$, $\texttt{min\_samples} \in (0, 50)$ in steps of $5$ for the MNIST dataset. 

For OPTICS $\texttt{max\_eps} \in (0.1, 0.5)$ in steps of $0.025$, $\texttt{min\_samples} \in (0, 50)$ in steps of $5$ for IRIS; $\texttt{max\_eps} \in (0.1, 0.5)$ in steps of $0.025$, $\texttt{minPts} \in (0, 50)$ in steps of $5$ for Banknote; and $\texttt{max\_eps} \in (0.1, 3)$ in steps of $0.025$, $\texttt{min\_samples} \in (0, 50)$ in steps of $5$ for the MNIST dataset. 

The runtimes for the best ARI scores achieved are provided in Table \ref{tab:reduceddimensiondata}. The exact hyperparameters and code used to generate the plots are given in the Github repository.

Consistent with the observations made in the preceding section, DBSCAN exhibits substantially lower runtimes than CLASSIX\_M on small-scale datasets such as Iris and Banknote. However, this advantage diminishes as the dataset size increases. In particular, on large-scale datasets such as MNIST, DBSCAN demonstrates worse scalability, with CLASSIX\_M showing significantly reduced runtime.

With respect to clustering quality, dimensionality reduction via UMAP appears to negatively impact performance on the Banknote dataset across all considered algorithms, as evidenced by consistently lower ARI scores. Nevertheless, CLASSIX\_M achieves substantially higher clustering accuracy than its competitors. Finally, OPTICS is uniformly outperformed by both CLASSIX\_M and DBSCAN, in terms of both computational efficiency and clustering fidelity across all evaluated datasets.

\begin{table}[t]
\centering
\resizebox{\columnwidth}{!}{%
\begin{tabular}{l c c c c c c} 
\toprule
Dataset 
& \multicolumn{2}{c}{DBSCAN} 
& \multicolumn{2}{c}{OPTICS} 
& \multicolumn{2}{c}{CLASSIX} \\
\cmidrule(lr){2-3} \cmidrule(lr){4-5} \cmidrule(lr){6-7}
& ARI & Runtime 
& ARI & Runtime 
& ARI & Runtime \\
\midrule
IRIS & \textbf{0.84} & \textbf{0.008} & 0.29 & 0.05 & \textbf{0.84} & 0.03 \\
Banknote & 0.06 & \textbf{0.006} & 0.02 & 0.43 & \textbf{0.42} & 0.019 \\
MNIST & \textbf{0.54} & 0.677 & 0.04 & 22.81 & \textbf{0.54} & \textbf{0.139} \\
\bottomrule
\end{tabular}%
}
\caption{Comparison of best achieved ARI and their corresponding runtimes of DBSCAN, OPTICS and CLASSIX (Manhattan norm).}
\label{tab:reduceddimensiondata}
\end{table}


\begin{figure}[htbp]
  \centering
  \includegraphics[width=\columnwidth]{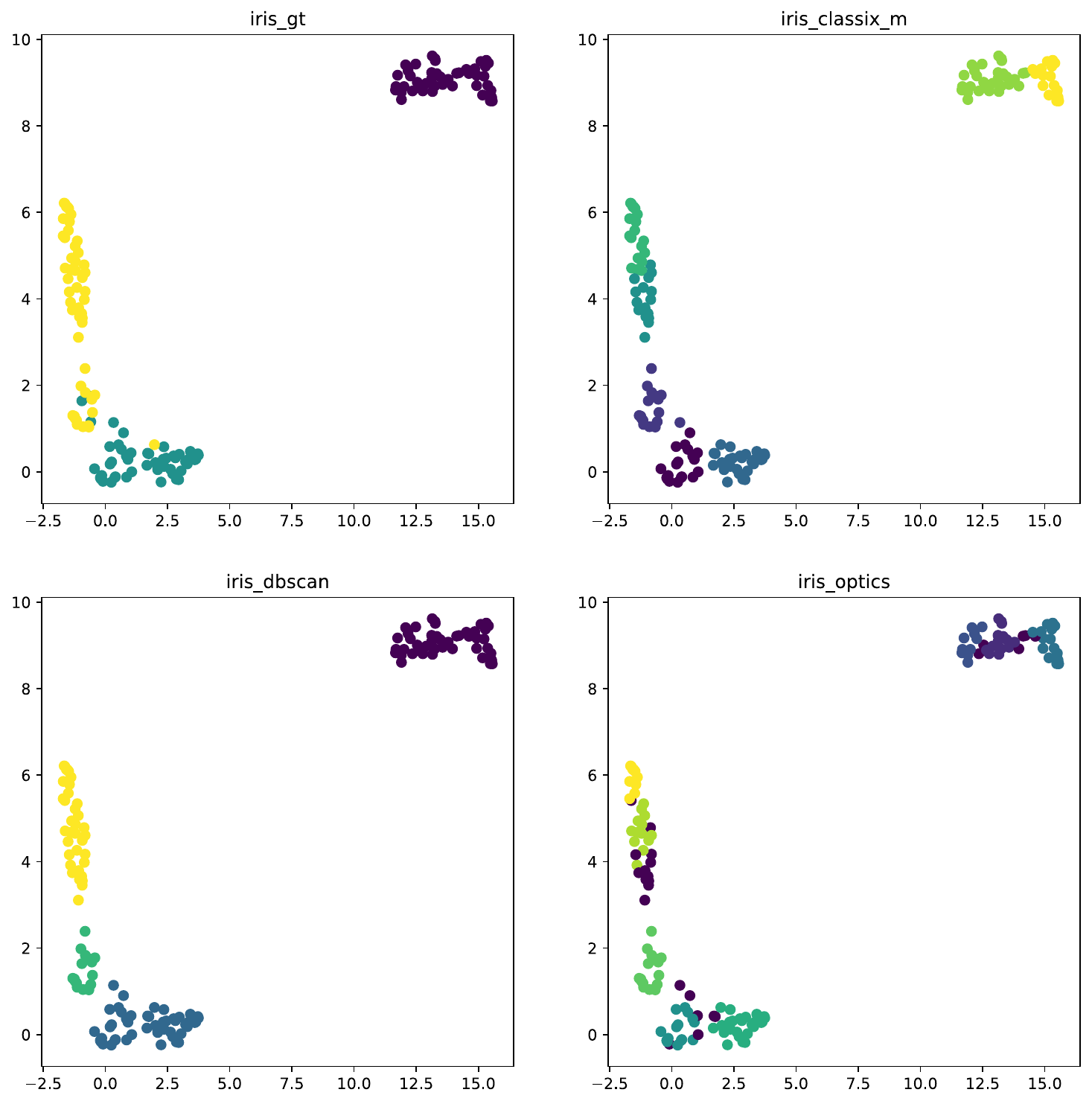}
  \caption{Performance comparison of DBSCAN, OPTICS, and CLASSIX Manhattan on the IRIS dataset}
  \label{fig:5}
\end{figure}

\begin{figure}[htbp]
  \centering
 \includegraphics[width=\columnwidth]{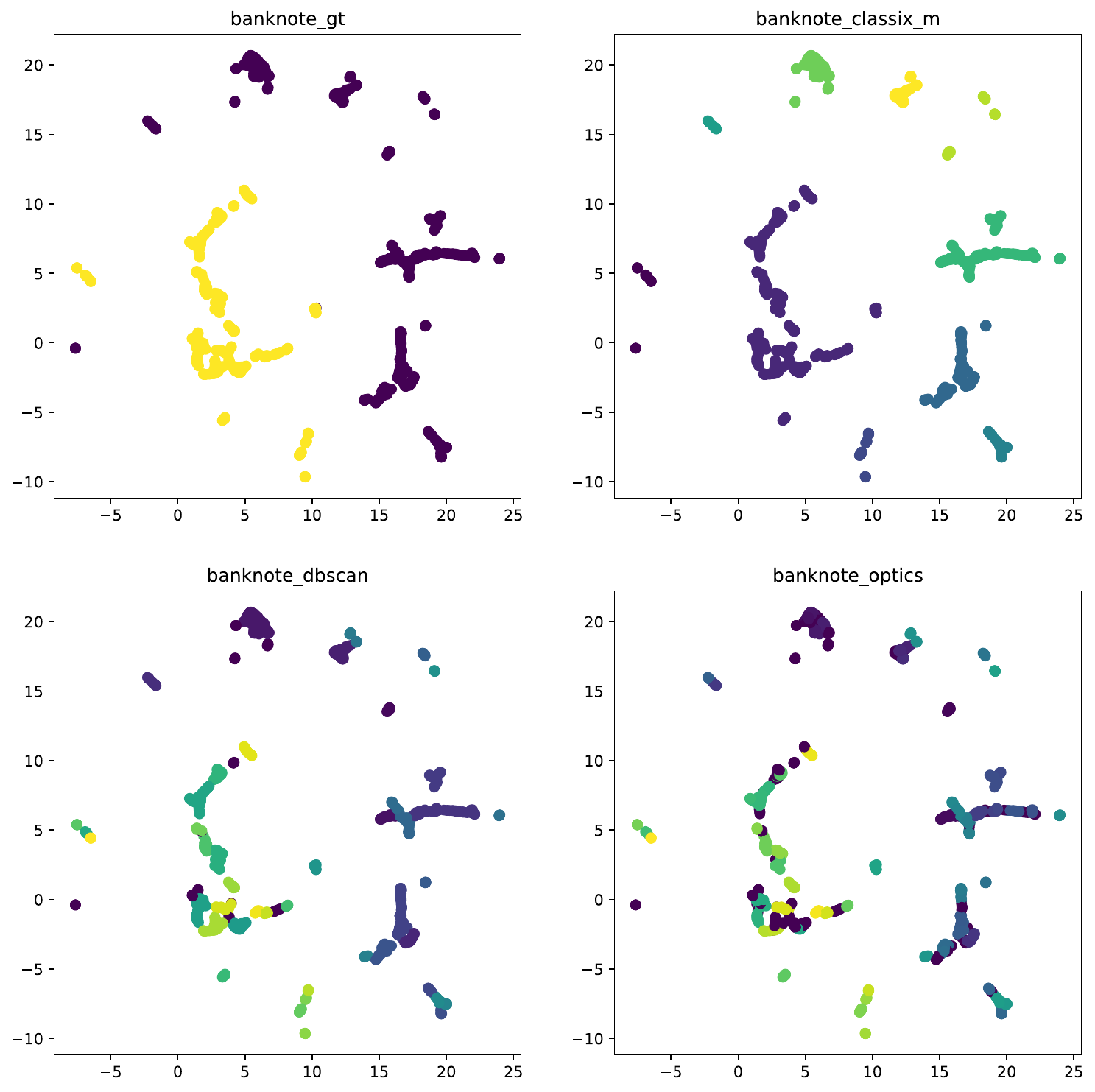}
  \caption{Performance comparison of DBSCAN, OPTICS, and CLASSIX Manhattan on the Banknote dataset}
  \label{fig:6}
\end{figure}


\begin{figure}[htbp]
  \centering
  \includegraphics[width=\columnwidth]{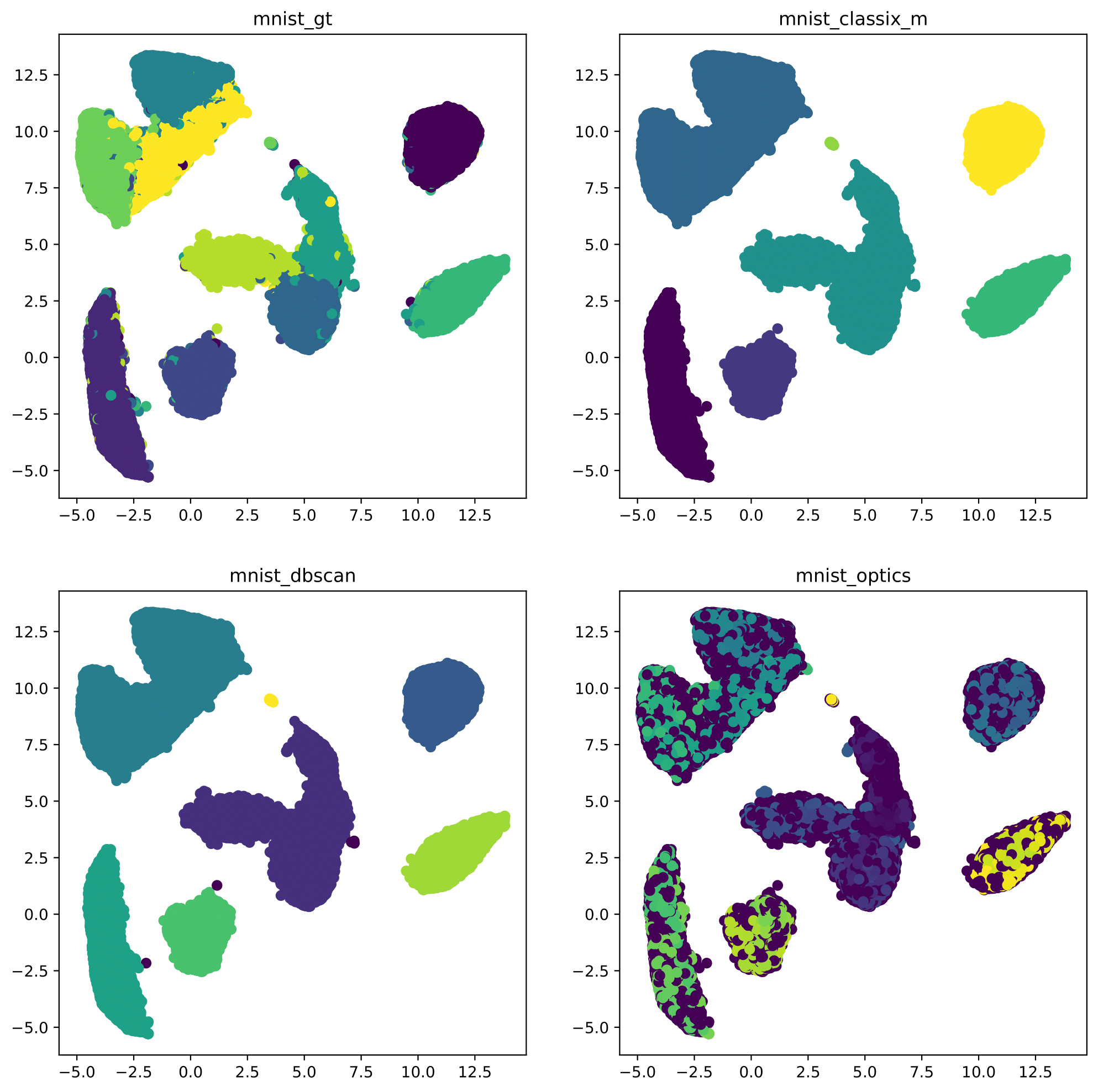}
  \caption{Performance comparison of DBSCAN, OPTICS, and CLASSIX Manhattan on the MNIST dataset}
  \label{fig:7}
\end{figure}

\end{subsection}

\end{section}

\section{Explainability}

Just like the original CLASSIX method, the CLASSIX Tanimoto or Manhattan versions employ  deterministic aggregation and merging stages, producing clusters that are fully reproducible and amenable to post hoc explanation.

In the aggregation stage, data points are aggregated into groups. This assignment induces an undirected graph on the vertex set $\{1,\dots,n\}$, represented by an adjacency matrix
\[
A^{(1)} \in \{0,1\}^{n \times n},
\]
where
\[
A^{(1)}_{ij} = 1 \quad \text{if} \quad x_i \text{ and } x_j \text{ are assigned to the same group},
\]
and $A^{(1)}_{ij} = 0$ otherwise. By construction, $A^{(1)}$ is symmetric and encodes the groups produced by the aggregation phase.

In the merging phase, the aggregated groups are merged into final clusters. Merging operations are added to the adjacency matrix.

When the optional \texttt{minPts} outlier handling is enabled, groups belonging to clusters smaller than the threshold are redistributed to neighboring clusters. In this case, edges in $A^{(1)}$ corresponding to invalidated cluster merges are removed, and all reassignment operations are recorded in an additional adjacency matrix
\[
A^{(2)} \in \{0,1\}^{ n \times n},
\]
which exclusively captures connections arising from \texttt{minPts}-based merging.

For explanatory queries, the adjacency information from the relevant stages is combined into a single graph representation. Given two data points $x_i$ and $x_j$, the CLASSIX explain function determines whether the points belong to the same final cluster and, if they do, then performs a breadth-first search on the union graph induced by $A^{(1)}$ and $A^{(2)}$. The algorithm then returns a concise summary of the sequence of aggregation and merging steps connecting them.

Crucially, the explicit separation of adjacency structures allows one to distinguish between connections arising from normal group aggregation and merging, and those induced by \texttt{minPts}-based reassignment. This transparency allows debugging and hyperparameter optimization, particularly in large-scale settings where exhaustive grid search is computationally infeasible.


\begin{section}{Conclusion}\label{conclusion}
In this work, we have demonstrated that the CLASSIX algorithm can be seamlessly adapted to different distance metrics, and discussed a general strategy for extending it to any norm. We have also demonstrated that the algorithm can be further optimized by incorporating sharper search termination inequalities where applicable. 
Our approach stands out from currently popular clustering methods on two fronts. Firstly, our novel search termination approach using scoring greatly reduces the number of pairwise distance computations, thereby reducing computational load. Secondly, the simple two-step non-iterative aggregation and merging stages are completely deterministic and allow greater explainability in the clusters generated, making further analysis easier.

Since CLASSIX does not require the computation and storage of the full pairwise distance matrix, it can run on much lower memory machines. By contrast, the Taylor--Butina algorithm fails to run on a M1-Macbook Air with 8\,GB memory, throwing an out-of-memory exception after a few minutes on a dataset of 91,000 molecules. CLASSIX\_T, on the other hand, completes the entire clustering within a minute, including computation of the relevant pairwise distances. CLASSIX\_T could easily be sped up even more by using bitwise operations (our current implementation uses integer operations).

For CLASSIX\_M, the probabilistic analysis that we presented is based on a simple model of uniformly distributed data points. A more realistic model could be based, e.g., on Gaussian-distributed clusters. We have left this for future work. An interesting further optimization approach of both CLASSIX\_M and CLASSIX\_T would be to incorporate GPU-support for the highly parallelizable distance computation and Boolean masking (cutoff condition checking) steps, which would improve performance on larger datasets.

\end{section}

\sloppy
\section{Acknowledgments} We are grateful to  Jacob Gora who made us aware of the Tanimoto distance clustering problem. S.\,G.\ acknowledges funding from the UK’s Engineering and Physical
Sciences Research Council (EPSRC grant EP/Z533786/1) and the Royal Society (RS
Industry Fellowship IF/R1/231032). K.\,R.\  acknowledges a partial PhD scholarship from The MathWorks. 

\bibliographystyle{elsarticle-num}

\bibliography{refs} 

\end{document}